# BCFPL: Binary classification ConvNet based Fast Parking space recognition with Low resolution image

Shuo Zhang[a], Xin Chen*[a], Zixuan Wang[a]
[a]Center for applied mathematics, Tianjin University, Tianjin, 300072, China.
*Corresponding author: chen_xin@tju.edu.cn

**ABSTRACT**

The automobile plays an important role in the economic activities of mankind, especially in the metropolis. Under the circumstances, the demand of quick search for available parking spaces has become a major concern for the automobile drivers. Meanwhile, the public sense of privacy is also awaking, the image-based parking space recognition methods lack the attention of privacy protection. In this paper, we proposed a binary convolutional neural network with lightweight design structure named BCFPL, which can be used to train with low-resolution parking space images and offer a reasonable recognition result. The images of parking space were collected from various complex environments, including different weather, occlusion conditions, and various camera angles. We conducted the training and testing progresses among different datasets and partial subsets. The experimental results show that the accuracy of BCFPL does not decrease compared with the original resolution image directly, and can reach the average level of the existing mainstream method. BCFPL also has low hardware requirements and fast recognition speed while meeting the privacy requirements, so it has application potential in intelligent city construction and automatic driving field.

## 1. INTRODUCTION

With the rapid increase of urban residents, parking difficulties have become more and more severe. To find a parking space would be a regular requirement which may waste a lot of time for all the drivers in the city. Non-image-based parking space recognition methods usually use geomagnetic sensors[1] to detect parking status, but this method has complicated installation procedures, difficult follow-up maintenance and high cost. Meanwhile, there are radio frequency[2], ultrasonic[3], radar[4] and other methods to detect parking Spaces, but these methods generally have the disadvantages of high cost and poor universality. Compare with these methods, the image-based method determines the parking status through the visual algorithm, which not only has low cost and high accuracy, but also can detect a wide range of parking status, which is very suitable for the scene of open parking lot.

Image-based parking space detection method has become a general solution. More and more affect has been made to improve the accuracy and efficiency of the algorithm. In 2002, Dan et al.[5] used Support Vector Machine (SVM) classifier to extract color vector features to distinguish parking areas from non-parking areas. In 2007 Wu et al.[6] attempted to overcome the occlusion problem of this approach by classifying the states of three adjacent spaces as a unit and defining the color histograms spanning the three spaces as features in the SVM classifier. In 2013, Huang, Tai and Wang[7] constructed a vacant parking space detection system based on three-dimensional parking space model using Bayesian hierarchical framework. In 2014, Jermsurawong, Ahsan, Haidar, Haiwei and Mavridis[8] used specially trained customized neural networks to determine the parking space occupancy state and parking demand according to the visual features extracted from parking Spaces. In 2015, De Almeida et al.[9] trained multiple texture features of SVM classifiers, such as LBP, LPQ and their variations. They also used the integration of SVM to improve detection performance, applying simple aggregation functions such as maximum or average values to the confidence values given by the classifier.

With the rapid development of deep learning area, convolutional neural network models have been widely applied to parking space recognition. In 2016, Giuseppe Amato et al.[10] constructed mAlexNet, a binary classification network model, with reference to AlexNet[11], which greatly improved the accuracy of recognition, the accuracy rate can reach 90% on some datasets. In 2020, Shen et al.[12] integrated non-local operations to gauge pixel similarity across distances, extracting high-frequency edge features tailored for parking slot images. This enhancement significantly boosted recognition accuracy, surpassing 95% on the designated dataset. Fukusaki et al[13] proposed a parking detection method using CNNs. They trained the networks across various ambient light conditions to effectively handle challenges associated with light changes. In 2023, Satyanath G et al[14]. proposed a method for detecting parking Spaces in hazy weather using CNN, which improves the accuracy of parking space detection in such weather conditions. Although the identification of parking space image has reached a high accuracy, these methods do not consider the protection of privacy. Meanwhile, the detection efficiency of these methods is not high.

In recent years, big data has evolved from being merely data to a crucial production factor, profoundly influencing various industries and reshaping societal activities. While facilitating data-related processes, such as generation, collection, storage, transmission, and utilization, it also raises concerns about the growing threat to public privacy[15]. The widespread use of information technology amplifies the risk of privacy data exposure. Installing cameras in parking lots for parking identification poses a significant risk of information leakage. Certain cameras can capture detailed information, including license plate numbers and even the faces of drivers. This data may be exploited by malicious actors for analyzing travel patterns, selling personal information, engaging in targeted fraud, and intrusive advertising. Opting for the direct acquisition and use of low-resolution images is an effective measure to safeguard privacy. As shown in Figure 1, with the decrease of resolution, parking space related information becomes less and less obvious. In the past, image-based parking detection methods did not take privacy protection into account. To the best of our knowledge, our job is the first work that use low-resolution images for parking space recognition. Low-resolution parking image has less data and more prominent features, so it is suitable for binary classification task. At the same time, we designed a small convolutional neural network BCFPL, which is suitable for low resolution parking image recognition. BCFPL reduces the hardware requirements, has more advantages in recognition speed and image transmission, so it has high great potential in the areas such as smart city and autonomous driving.

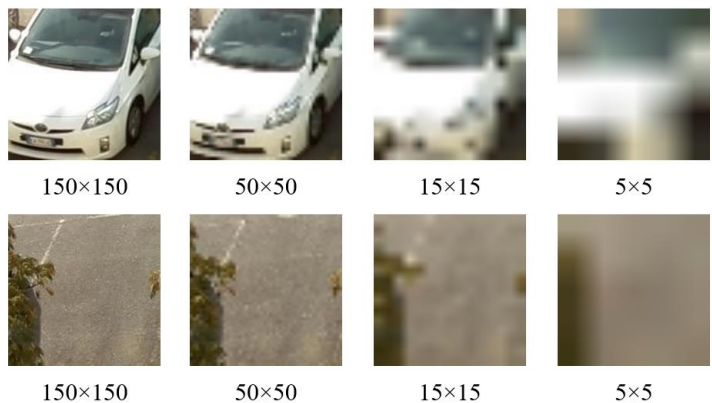

Figure 1. Images with different resolutions showing whether the parking space is occupied or not.

## 2. DATASTES

This paper employs three parking space datasets. PKLot[8] and CNRPark-Ext[9] are publicly accessible datasets, whereas the third dataset, named MiniPK due to its limited image count, was specifically compiled for this study.

PKLot dataset contains 695,899 images of parking spaces, which were divided into three subsets: UFPR04, UFPR05 and PUCPR. All images were taken and acquired at UFPR and PUCPR outdoor parking lots under different weather conditions such as sunny, rainy and cloudy weather.

CNRPark-Ext dataset contains 144965 images of parking spaces, which were acquired by 9 cameras in different weather environments. The occlusion conditions are stricter, including partial or even total occlusion, such as street lights and leaves. CNRPark-Ext contains images with lower brightness, which is helpful for the model to cope with the reality that the camera capture image is not clear when the sky is dark. The data set also provided CNRPark, which was a subset of 12,584 parking space images that distinguished between odd and even parking spaces.

MiniPK dataset contains 100 images of parking spaces, which are obtained separately from more than 20 images of open parking lots, among which 50 parking spaces are empty and 50 parking spaces are occupied. When collecting images for MiniPK, we took shots from different angles. The Environmental factors such as light, weather, and occlusion state are more complex and variable than other datasets we know of. For each image of the parking lot, we only take no more than five images of parking spaces to avoid the problem that the images of the same parking lot are too similar, so the richness is strong. Therefore, the dataset is ideal for testing the generalization of models.

## 3. METHODS

### 3.1 Image resolution reduction

One of the key points of this work is to explore the relationship between different resolution images and the recognition accuracy of parking spaces. The experiments need to be carried out on different level of low-resolution images, but the existing datasets do not meet the conditions of low resolution, so it is necessary to reduce the resolution of images to obtain low-resolution parking image datasets.

Given the imperative to diminish the resolution of parking space images to bolster privacy protection, there is currently a lack of universally established guidelines in this regard. To address this, we undertook a comprehensive analysis of the dataset's image sizes, revealing that the original images generally exceeded 50×50 pixels. Consequently, we initiated our experimentation with a 50×50 resolution. In each subsequent step, we reduced both the length and width of the image by approximately 30%. Simultaneously, with each reduction, the number of pixels in the image was roughly halved. This led to a sequence of resolution groups: 50×50, 35×35, 25×25, 18×18, 13×13, 9×9, 7×7, 5×5, 3×3, comprising a total of nine groups. While striving to maintain recognition accuracy, we observed that lower resolutions provided enhanced privacy protection.

The resolution reduction process is performed as follows. Let $a$ and $b$ be the length and the width of an original image, let $m$ and $n$ be the length and the width of the scaled image. $Q$ is one pixel in scaled image where $i$ and $j$ are the coordinate value of $Q$. Then the pixel in $(i, j)$ position of the scaled image correspond to $(j \times a/m, i \times b/n)$ position of the original image. For convenience, we write this point as $(x_0, y_0)$. Since this position may be a floating-point number (such as point $P$ in Figure 2, it is necessary to find the position of four nearby pixels in the original image and calculate the pixel value of this point. The bilinear interpolation method is used to calculate the pixel value.

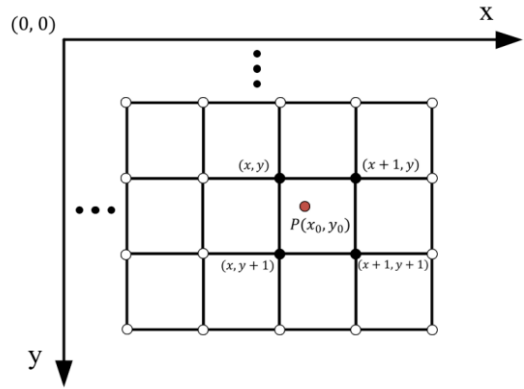

Figure 2. The correspondence of a point in the scaled image to the original image.

As $P$ and $Q$ are corresponding points, calculating the pixel value at point $Q$ is simply a matter of computing the pixel value at point $P$. Let x and y be the values obtained by rounding down the coordinate values of $P$. Then we get four points around $P$: $(x, y)$, $(x+1, y)$, $(x, y+1)$, $(x+1, y+1)$. The corresponding pixel values are respectively $f(x, y)$, $f(x+1, y)$, $f(x, y+1)$, $f(x+1, y+1)$. $Z_1$ and $Z_2$ are the interpolation in $x$ direction. We use Formula (1) and (2) to calculate $Z_1$ and $Z_2$.

$$Z_1 = (f(x+1, y) - f(x, y)) \times (x_0 - x) + f(x, y) \quad (1)$$

$$Z_2 = (f(x+1, y+1) - f(x, y+1)) \times (x_0 - x) + f(x, y+1) \quad (2)$$

Z is the pixel interpolation in $y$ direction. We use Formula (3) to calculate $Z$.

$$Z = (Z_2 - Z_1) \times (y_0 - y) + Z_1 \quad (3)$$

The obtained pixel value $Z$ is regarded the pixel value of point $P$, which corresponds to the pixel value of the point $Q$ in the scaled image. Since $Q$ is an arbitrary point, by bilinear interpolation algorithm, the pixel value of each point in the scaled image can be calculated.

### 3.2 CNN for parking space detection

To enhance detection efficiency, we devised a CNN model featuring two convolutional layers and two fully connected layers. Employing established datasets, this model is employed for binary classification of parking space images. Detailed information regarding the model's architecture is depicted in Figure 3. The two convolution layers within the model employ 7×7 convolution kernels to extract local features. In the first convolution layer, 8 convolution kernels are utilized for feature extraction, with the convolution kernels employing a sliding step of 3. In the second convolution layer, 16 convolution kernels are employed for feature extraction, and the sliding step is set to 2. Following each convolution layer, the nonlinear activation function ReLU is applied, and the distribution is subsequently smoothed through batch normalization operations. Following the two convolutional layers, we incorporated two fully connected layers into the model. The first fully connected layer consists of 60 neurons, while the second one comprises 2 neurons. To expedite model convergence and mitigate overfitting, we incorporated a 50% dropout operation within the first fully connected layer. The output of the second fully connected layer serves as the classification criterion.

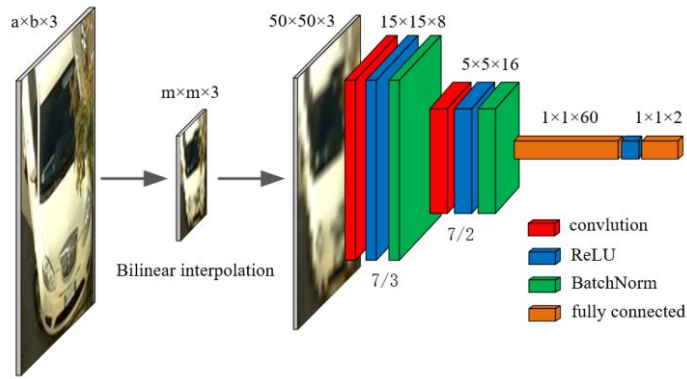

Figure 3. Image processing flow and CNN structure: the parameters of the feature map are described as "length × width × channel num". For convolutional layers, parameters are specified as "size/stride num".

The model's input size was configured to 50×50. To ensure compatibility with the input size of the binary classification network, we meticulously adjusted the dimensions of images with varying resolutions subsequent to resolution reduction. Throughout this process, we consistently employed the bilinear interpolation.

It's essential to highlight that the act of increasing the image resolution following a reduction is an irreversible process. The features that were lost during the initial resolution reduction cannot be fully regained during resolution enhancement. Consequently, an image with reduced resolution can provide an approximate representation of what would be captured by a low-resolution camera in a real-world scenario.

## 4. EXPERIMENTS

All experiments were based on the PyTorch framework, with the CPU of Intel(R) Core(TM) i5-1135G7 @ 2.40GHz and an NVIDIA GeForce MX450 GPU. The OpenCV library was used in making parking space masks and adjusting image resolution. We used the images with reduced resolution in the datasets to carry out the experiment. PKLot and CNRPark-Ext were used for training and testing, and MiniPK was only used to verify the generalization of the model.

During training phase, we used horizontal flipping technique of the training images for data augmentation: images are enlarged to a resolution of 50×50 and then are horizontally flipped with a probability of 0.5 as input to the neural network. During prediction, images are only resized to 50×50 without flipping.

We selected AdamW as the optimizer for training, cross-entropy loss function as the loss function. The models were trained 20 epochs with a learning rate of 0.001 halved every 4 epochs, a batch size of 128. In the following experiments, the above parameters were used except for the smaller learning rate and more training epochs when exploring the overfitting problem in 4.2.

**4.1 Recognition effect with different resolution**

We employed our proposed convolutional model for both training and testing. In the case of the CNRPark-Ext dataset, 2,000 images were randomly allocated for training, while an additional 10,000 images were set aside for testing. Similarly, for the PKLot dataset, 10,000 images were randomly selected for model generalization testing, and 100 images were chosen from the MiniPK dataset for the same purpose. Subsequently, we substituted the training dataset with PKLot and repeated the experimental procedure to observe accuracy changes across datasets with varying resolutions. Detailed results are illustrated in Figure 4.

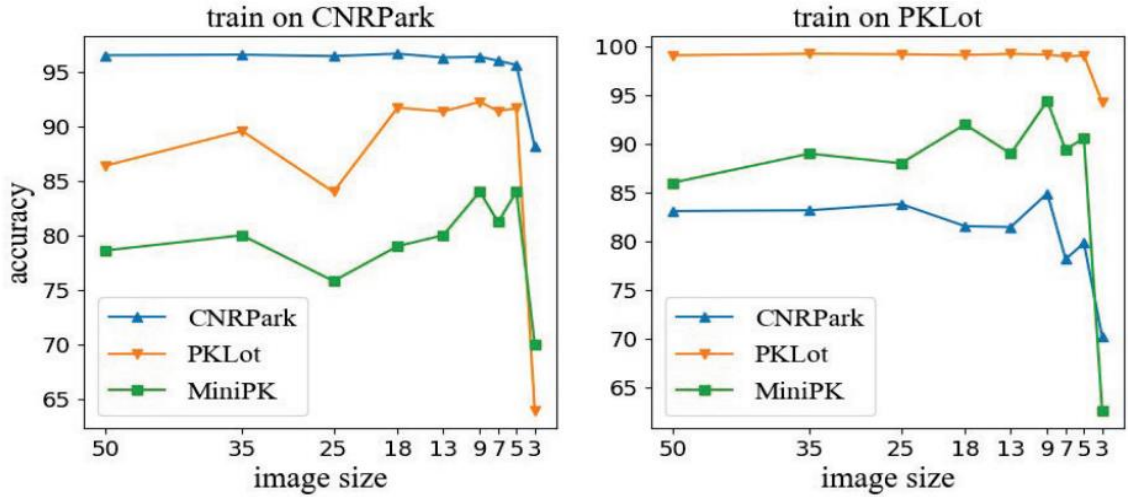

Figure 4. Training on CNRPark and PKLot, the relationship between accuracy and image resolution of different datasets. (The horizontal coordinate is the length or width of the image).

When the image resolution was reduced for the experiment, the test accuracy of the model did not decrease within a certain resolution range, and the test results were still around 90% until the image size was reduced to 7×7. Moreover, in some experimental datasets, the test accuracy of low-resolution images is even higher. When the resolution dropped below 5×5, the test accuracy of the model decreases significantly, indicating that the parking space image with too few pixels did not have enough significant features for the model to extract. Therefore, it is feasible to reduce the image resolution in a certain range for the task of parking space recognition.

**4.2 Explore the overfitting problems**

The overfitting of the model is an important reason for the low accuracy of the test results, especially in the higher resolution. Although we adopted some procedures to reduce overfitting in data processing and model design, the phenomenon of overfitting still exists. In order to explore the overfitting problem in the training process, we randomly selected 2,000 images in CNRPark-Ext as the training set and trained them on the two-layer convolution model. At the same time, we used a smaller learning rate of $2\times10^{-5}$ without attenuation and increased the number of training epochs. After the completion of each training epoch, a test is conducted.

In experiments using images with higher resolution, the accuracy rate on PKLot quickly peaked and began to decline steadily. The results are shown in Figure 5. As resolution de-creases, the decline in accuracy begins to abate. When the resolution is reduced to about 9×9, the accuracy of the test set remains stable, and the phenomenon of accuracy reduction no longer occurs. However, when the resolution continues to decrease to 5×5 or below, the accuracy is always at a low level. On MiniPK dataset, the accuracy of the test was higher at low-resolution.

For higher resolution images, the training process is unstable and have obvious overfitting problems. With the above parameters, the model obtained by training 30 epochs is optimal. For low-resolution images in the range of 9×9 to 7×7, the accuracy of the test set remained at a high level, and the accuracy did not decrease with the increase of the number of training rounds. Low resolution images are more stable for the training process, and increasing the number of training rounds will not significantly reduce the accuracy. Therefore, using low-resolution images within a certain range is more friendly to the training process while meeting the task requirements.

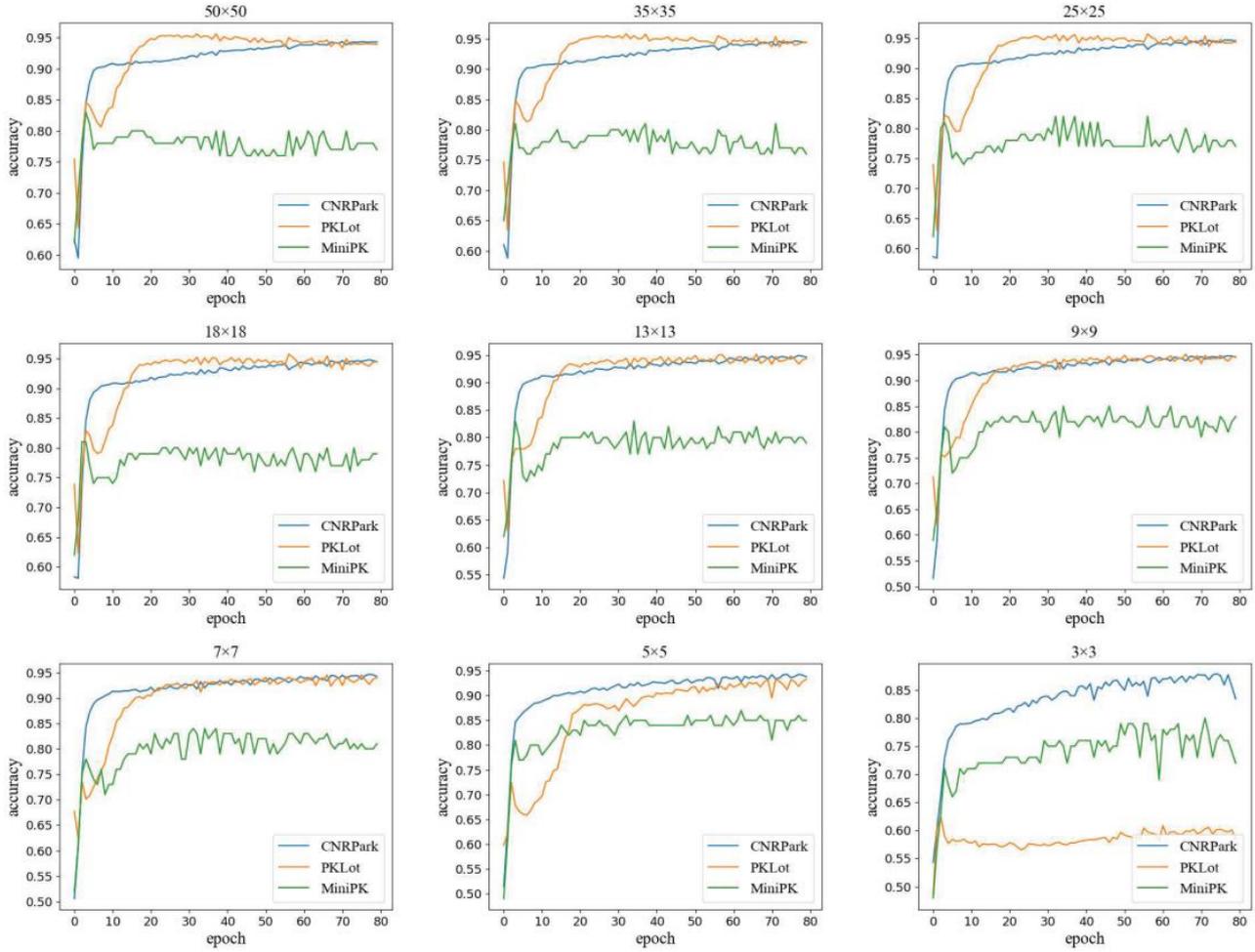

Figure 5. Training with different resolution images, the test accuracy varies with the number of training epochs.

## 4.3 Evaluations

With the best parameters of our model, further generalization tests were conducted between subsets of CNRPark. CNRPark is divided into even and odd parking Spaces (CNRPark Even and CNRPark Odd). Training was done on one of the subset and testing was done on the other, all images in dataset were used in training and testing. In both the training and testing phases, we used images of parking spaces with a resolution of 7×7.

We compare our approach with that of Giuseppe Amato[10], De Almeida et al[9]. mAlexNet proposed by Giuseppe Amato is a deep learning model, which has a convolutional neural network structure with three convolutional layers and two fully connected layers. The input size of mAlexNet is 224×224. De Almeida used a parking occupancy detection method based on texture feature histogram training of RBF kernel support vector machine, which is a relatively advanced method of non-deep learning. The indexes we compared are accuracy rate and AUC(Area Under Curve). The ROC(Receiver Operating Characteristic) curve shows the change of true rate with false positive rate, and AUC is defined as the area under the ROC curve enclosed by the axes. AUC measures the slope degree of the curve near the perfect classification point (0,1), and the value range is 0-1. When the AUC is 0.5, the classification is completely random, and when the AUC is 1, the classification is completely correct. The results are shown in Table 1.

From the experimental results, the deep learning method is obviously better than the non-deep learning method. Compared with mAlexNet, BCFPL demonstrates an enhanced accuracy of 1% to 1.5%.

Table 1. Training on one subset of CNRPark and testing the accuracy on the other subset.

| Train set | Test set | Method | Accuracy | AUC |
|---|---|---|---|---|
| CNRParkOdd | CNRParkEven | BCFPL | **91.61%** | **0.97** |
|  |  | mAlexNet | 90.13% | 0.94 |
|  |  | LPQgd | 87.65% | 0.95 |
| CNRParkEven | CNRParkOdd | BCFPL | **91.75%** | **0.97** |
|  |  | mAlexNet | 90.71% | 0.92 |
|  |  | LBP | 87.21% | 0.92 |

In order to further demonstrate the superiority of BCFPL, we compared the training time and test time with mAlexNet. The experiments were all conducted on the same equipment and with the same parameter configuration. The time spent is shown in Table 2.

Table 2. Compare the time spent by BCFPL and mAlexNet when training and testing with different numbers of images.

| Mode | Image num | Method | Time |
|---|---|---|---|
| Train | 10000 | BCFPL | 1m36s |
|  |  | mAlexNet | 4m27s |
|  | 20000 | BCFPL | 3m14s |
|  |  | mAlexNet | 9m3s |
| Test | 100000 | BCFPL | 43s |
|  |  | mAlexNet | 2m20s |
|  | 200000 | BCFPL | 1m23s |
|  |  | mAlexNet | 4m38s |

With minimal parameters and low-resolution images, BCFPL demonstrates a threefold speed improvement compared to mAlexNet in training and testing times. This efficiency is attributed to smaller input images and a simpler model structure. In practical scenarios, where parking space images are transmitted to servers for identification, our low-resolution approach not only excels in speed but also in data storage and transmission efficiency. Notably, using a 7×7 resolution image reduces data volume to less than 1/50 compared to the original image's resolution of 50×50.

## 5. CONCLUSION

Reducing the resolution of parking space images within the range of 7×7 to 9×9 significantly enhances model practicality, maintaining test accuracy over 90%. This resolution range effectively mitigates overfitting issues, allowing for increased training rounds without compromising model generalization. The proposed BCFPL model, featuring two convolutional and two fully connected layers, simplifies the identification of parking spaces, exhibiting strong generalization with minimal parameters (30,000). Our approach meets the need for privacy, and at the same time, requires less hardware, costs less, and is faster. It has great application prospects in smart cities, autonomous driving and other fields in the future.

## REFERENCES

[1] Liu G, He W. Research on Geomagnetic Differential Detection Algorithm of Roadside Parking Vehicles[C]//2022 4th International Conference on Machine Learning, Big Data and Business Intelligence (MLBDBI). IEEE, 2022: 345-349.
[2] Atiqur R. Radio frequency identification based smart parking system using internet of things[J]. IAES International Journal of Robotics and Automation, 2021, 10(1): 41.
[3] Allbadi Y, Shehab J N, Jasim M M. The smart parking system using ultrasonic control sensors[C]//IOP Conference Series: Materials Science and Engineering. IOP Publishing, 2021, 1076(1): 012064.

[4] Li J, Chen H, Yang Q, et al. LiDAR-Based High-Accuracy Parking Slot Search, Detection, and Tracking[R]. SAE Technical Paper, 2020.
[5] DAN N. Parking management system and method[P]. US,20030144890, 2003.
[6] Wu, Q., Huang, C., Wang, S.-y., Chiu, W.-C., & Chen, T. (2007). Robust parking space detection considering inter-space correlation. In Multimedia and expo, 2007 ieee international conference on (pp. 659–662). IEEE.
[7] Huang, C.-C., Tai, Y.-S., & Wang, S.-J. (2013). Vacant parking space detection based on plane-based bayesian hierarchical framework. Circuits and Systems for Video Technology, IEEE Transactions on, 23, 1598–1610.
[8] Jermsurawong, J., Ahsan, U., Haidar, A., Haiwei, D., & Mavridis, N. (2014). One-day long statistical analysis of parking demand by using single-camera vacancy detection. Journal of Transportation Systems Engineering and Information Technology,14, 33–44.
[9] de Almeida, P. R., Oliveira, L. S., Britto, A. S., Silva, E. J., & Koerich, A. L. (2015).Pklot–a robust dataset for parking lot classification. Expert Systems with Applications, 42, 4937–4949.
[10] Amato G, Carrara F, Falchi F, et al. Deep learning for decentralized parking lot occupancy detection[J]. Expert Systems with Applications, 2017, 72: 327-334.
[11] Alex Krizhevsky, Ilya Sutskever, and Geoff Hinton. Imagenet classification with deep convolutional neural networks. In NeurIPS, 2012.
[12] SHEN X, SHEN Z, HUANG Y, et al. Deep convolutional neural network for parking space occupancy detection based on non-local operation[J]. Journal of Electronics & Information Technology, 2020, 42(9): 2269-2276.
[13] Fukusaki T, Tsutsui H, Ohgane T. An evaluation of a CNN-based parking detection system with Webcams[C]//2020 Asia-Pacific Signal and Information Processing Association Annual Summit and Conference (APSIPA ASC). IEEE, 2020: 1-4.
[14] Satyanath G, Sahoo J K, Roul R K. Smart parking space detection under hazy conditions using convolutional neural networks: a novel approach[J]. Multimedia Tools and Applications, 2023, 82(10): 15415-15438.
[15] Demirol D, Das R, Hanbay D. A key review on security and privacy of big data: issues, challenges, and future research directions[J]. Signal, Image and Video Processing, 2023, 17(4): 1335-1343.